\documentclass[conference]{IEEEtran}

\IEEEoverridecommandlockouts

\usepackage{amsmath,amssymb,amsfonts}
\usepackage{algorithmic}
\usepackage{graphicx}
\usepackage{textcomp}
\usepackage{xcolor}
\usepackage[backend=biber, style=numeric, sorting=none]{biblatex}
\usepackage{hyperref, url}
\usepackage{caption}
\usepackage{multirow}
\usepackage{booktabs}

\hypersetup{
  colorlinks   = true, 
  urlcolor     = blue, 
  linkcolor    = blue, 
  citecolor   = blue 
}
\def\BibTeX{{\rm B\kern-.05em{\sc i\kern-.025em b}\kern-.08em
    T\kern-.1667em\lower.7ex\hbox{E}\kern-.125emX}}

\begin{document}

\title{
Neural Activity Synchronization-Based Fast Local Learnable Spiking Neural Network Architecture
}

\author{Anonymous Submission}


\author{
\IEEEauthorblockN{
Pranith Dutta\textsuperscript{1}, Sk Wasim Rajjak\textsuperscript{1}, Siddhanta  Ray\textsuperscript{1}, Santlal Prajapati\textsuperscript{2}, Manobendra Nath Mondal\textsuperscript{2}}

\IEEEauthorblockA{\textsuperscript{1}\textit{Department of Computer Science and Engineering, Sister Nivedita University, Kolkata, India} \\ \textsuperscript{2}\textit{Department of Computer Science and Engineering, UPES Dehradun, Dehradun, India}\\  
\{pranithdutta26, wrazzak24, siddhanta2004\}@gmail.com\\santlal.prajapati@upes.ac.in, manobendra.mondal@ddn.upes.ac.in}
}


\maketitle

\begin{abstract}

Implementing intelligent learning systems on low-power edge devices is challenging due to the high computational complexity, training latency, and energy demands of traditional Artificial Neural Networks (ANNs). Spiking Neural Networks (SNNs), biologically plausible and spike-driven, come as an alternative. However, SNN training commonly relies on backpropagation through time (BPTT), which is computationally expensive, inherently global, and unsuitable for battery-driven devices. This work proposes a novel local learning framework for multi-layered SNNs inspired by Neural Activity Synchronization (NAS). Each layer updates its own weights using locally computed gradients. No gradient backpropagation is needed. This significantly reduces computational complexity, training latency, and energy consumption. The proposed SNN model is trained and evaluated on the MNIST dataset, achieving a classification accuracy of $97.78\%$. The model is also analyzed with Gaussian and salt-and-pepper noise, demonstrating robust performance across different noise levels.
\end{abstract}


\section{Introduction}

Artificial Intelligence (AI) is becoming important because of its wide range of applications. Battery-driven intelligent devices need to run smart tasks with limited computational power, memory, and training time~\cite{10.1109/COMST.2023.3302474}. Artificial Neural Networks (ANNs) drive most of AI applications. However, ANNs are typically trained by gradient backpropagation~\cite{Rumelhart1986}. In this methodology, the gradient signals propagate from the output layer towards the input layer through hidden layers demanding high training latency, memory, and computational power. These requirements make ANNs unsuitable for on-device learning and deployment in such resource-constrained environments~\cite{patterson2021carbonemissionslargeneural}.

In neuro-biological system, groups of neurons exhibit synchronized activity during learning, memory, and information processing~\cite{10.3389/fncom.2017.00098}. This coordinated activations of group of neurons responding to common stimuli during information processing and learning is called Neural Activity Synchronization (NAS)~\cite{10.3389/fncom.2017.00098, 10943331}. This phenomenon has inspired learning methods that rely on local gradient rather than global gradient propagation. This idea is explored in ANNs as Local Learning Synchronization (LLS)~\cite{10943331}.

Spiking Neural Networks (SNNs) are a promising alternative to ANNs. They process information using spikes making them efficient in terms of time and power which suits resource-constraint devices~\cite{Gerstner_Kistler_2002,10985795, MAASS19971659}. However, supervised training of SNNs remains challenging as it relies on Backpropagation-Through-Time (BPTT) which stores neurons states across many time-steps~\cite{10.3389/fnins.2016.00508}. The surrogate gradient propagates in BPTT from the output layer back to the previous layers across temporal states. This leads to high memory usage and heavy computation because neuron states must be retained over multiple time-steps restricting efficient learning of power constraint devices ~\cite{10097504}.

Several approaches have tried to reduce this cost. Supervised Spike-Timing-Dependent Plasticity (STDP) approaches, like CoLaNET \cite{kiselev2025colanetspikingneural} and CDNA-SNN \cite{10428047} use local, biologically inspired synaptic plasticity rules to avoid backpropagation. CoLaNET relies on anti-Hebbian and dopamine-modulated plasticity. But it introduces randomness into training and needs a complex architecture with several hyperparameters tuned through a costly genetic search.
CDNA-SNN groups neurons into class-specific assemblies. Its layer-wise training extends the overall time-steps and requires removing untrained "hypoactive" neurons after every layer, which may compromise performance if not managed carefully. Gradient-based supervised learning methods remain the dominant approach for deep SNN training. However, HM2-BP \cite{jin2019hybridmacromicrolevelbackpropagation} employs a hybrid macro and micro-level backpropagation to improve the performance but errors still propagate using approximated spike train derivatives requiring memory and computational burden of layer wise error propagation. ReL-PSP \cite{DBLP:journals/tnn/ZhangWWBAZMQCCL22} introduces a neuron model for stable spike-timing based backpropagation. However, it still depends on full backpropagation and faces exploding and dead neurons problems.
The BS4NN \cite{10.1007/s11063-021-10680-x} binarizes synaptic weights to shrink memory and compute costs but comes with accuracy cost and requires an auxiliary set of full-precision proxy weights during training. PSAC \cite{Nazari2025} combines unsupervised Power-STDP with a reinforcement learning signal. However, it needs a large network whose size grows with the dataset, and its "unsupervised" middle layer still depends on a reinforcement supervised signal from labels.

This work addresses above challenges by proposing an energy-efficient and locally trainable SNN architecture. Inspired by NAS, the proposed SNN architecture uses fixed basis matrices to compute layer-wise error signals locally, eliminating global error signal propagation through the network's layers. This distinguishes the proposed method from BPTT-based SNN architecture.
The proposed SNN model achieves a classification accuracy of $97.78\%$ on MNIST dataset. Gaussian and salt-and-pepper noises are used to analyze the performance of the SNN architecture at various noise levels. The scalability of the architecture is verified on Fashion-MNIST dataset.

\begin{figure*}[t]
    \centering
    \includegraphics[width=\linewidth]{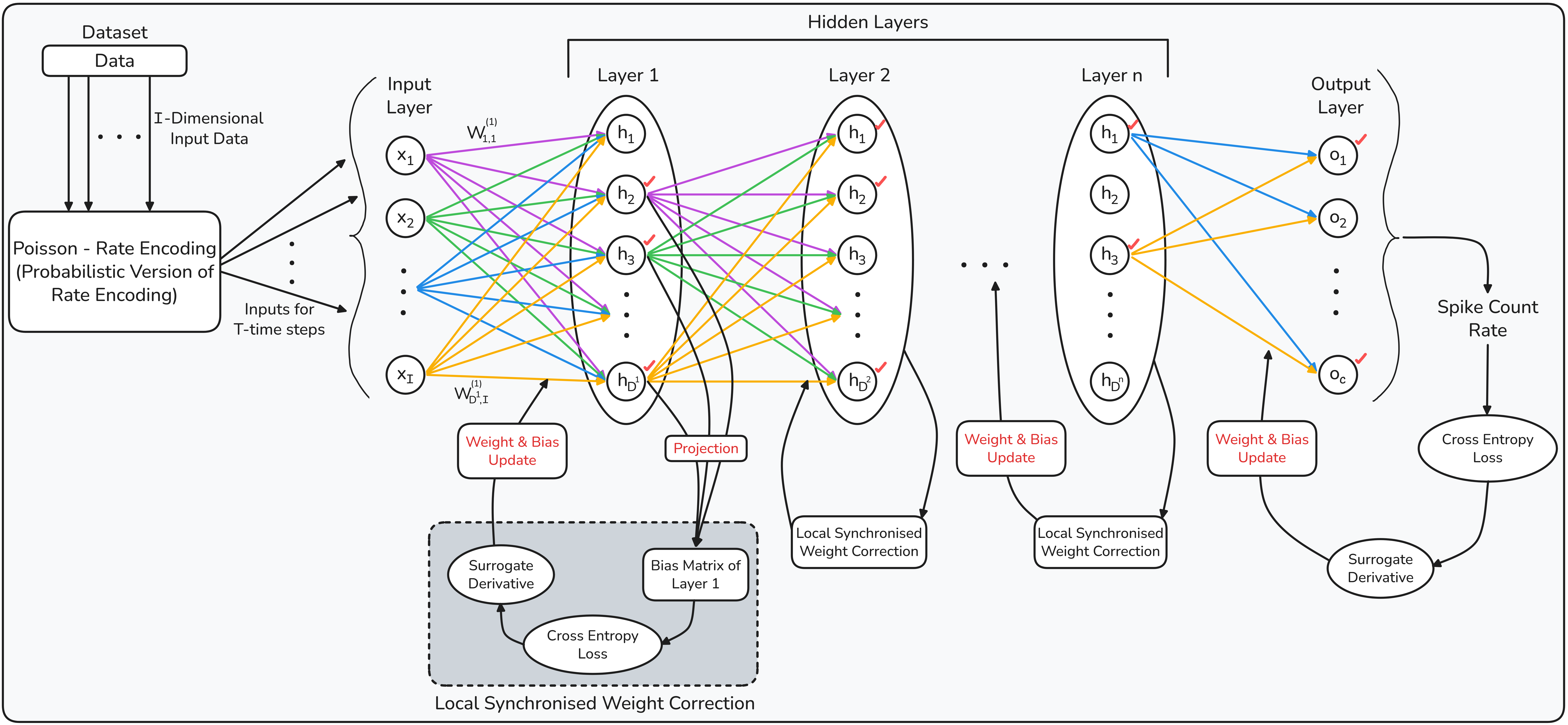}
    \caption{Neural Activity Synchronization based Spiking Neural Network architecture. This figure illustrates the processing of image data at a single time-step after Poisson rate encoding. The data is encoded for \(T\) time steps. 
    Here, \(I\) represents the input vector dimension, \(D^n\) denotes the number of neurons in the \(n\)-th hidden layer, and \(C\) represents the number of classes. The Red Tick mark indicates the firing of a neuron at that instant. 
    }
    \label{fig:model_arch}
\end{figure*}

Section \ref{back} covers the background and Section \ref{method} describes the proposed methodology. Section \ref{experiment} presents the experimental results, while Section \ref{conclusion} concludes the paper.

\section{Background}
\label{back}



\subsection{Spiking Neural Networks}
Spiking Neural Networks (SNNs) process information through discrete spikes rather than continuous activations, enabling sparse computation and low-power inference. SNNs have several neuron models including leaky-integrate-and-fire (LIF) and each offers a different trade-off between biological plausibility and computational performance \cite{Dutta2017}. Spike trains are generated by converting input data using encoding schemes like rate or temporal encoding \cite{Auge2021}. These allow static signals to be processed temporally within the SNNs \cite{9787485, 11201000}.



Each LIF neuron maintains its own membrane potential \(v_t\) at time-step \(t\). The synaptic current is added to the leaky contribution of the previous membrane potential, given by \(\alpha v_{t-1}\), where \(\alpha\) denotes the decay rate, to update the membrane potential \(v_t\).
When membrane potential crosses its threshold \(\vartheta\), the Heaviside step function \(H(\cdot)\) triggers spike \(s\) and membrane potential is then reset~\cite{Dutta2017}.

During training, the cross-entropy loss \(\mathcal{L}\) is computed at output layer using the logits and the target vector \(y\). The differentiation of \(\mathcal{L}\) with respect to the logits yields a class-level learning signal \(\delta\). Since the Heaviside function \(H(\cdot)\) is non-differentiable, gradient calculation is performed using fast-sigmoid derivative \(\phi'(\cdot)\) \cite{10.3389/fnins.2020.00119}. The synaptic weights are updated with weight gradient \(\Delta w\), computed as the negative of the product of the class-level learning signal, the surrogate derivative, and the previous synaptic spike activity,
along with learning rate \(\eta\) to control the magnitude of weight change \cite{8891809}.

\subsection{Local Learning and Synchronization}

Local learning refers to training paradigms in which synaptic updates rely on information locally available at each layer, avoiding global error propagation. Such approaches address key limitations of backpropagation, including update locking, high memory usage, and poor scalability to deep or resource-constrained systems.
Synchronization based learning is inspired by the synchronized activities of groups of neurons in response to common stimuli in biological systems~\cite{10943331}. Previous works have explored the use of fixed or structured signals to promote synchronized activities in neural networks~\cite{2024arXiv240805098K}. This setup allows efficient learning without global gradient flow~\cite{10943331}. These motivate the development of local learning mechanisms that combine locality with synchronized activity for energy-efficient and hardware friendly learning systems.

\section{METHODOLOGY}
\label{method}

This work proposes local learning SNN architecture by integrating Neural Activity Synchronization (NAS) into multi-layered SNNs. For a hidden layer, a fixed basis matrix is employed to project its spike-count activity into a local logit vector. This gives its own class-comparable error signal without requiring a signal propagated back from later layers.
In this section, the architecture overview, basis-matrix logit computation, and the local weight-update rule are described.

\begin{figure*}[t]
    \centering
    \includegraphics[width=\linewidth]{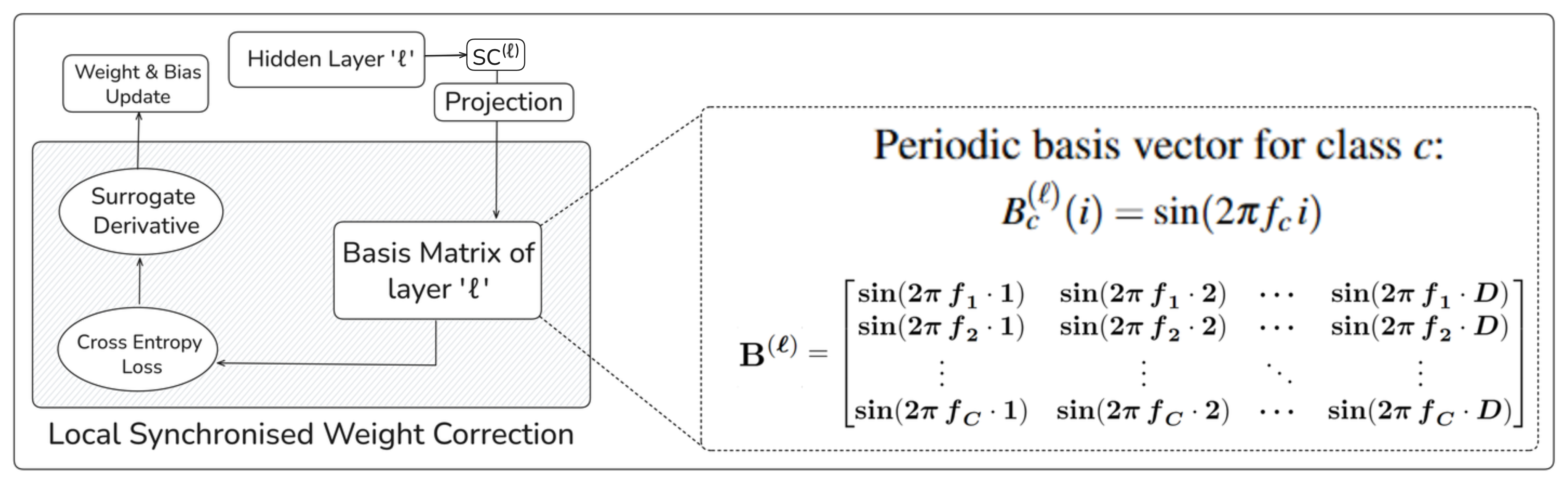}
    \caption{Implementation of the Local Neuronal Synchronous Model. The figure illustrates the formation of the basis matrix using periodic function in the proposed architecture. A distinct spatial frequency \(f_c\) is assigned to each class \(c\), thus they become class-specific periodic basis vectors. For a layer containing \(D\) neurons and \(C\) classes, the basis matrix $B \in \mathbb{R}^{C \times D}$ is formed by stacking these basis vectors. The spike count vector \(SC^{(\ell)}\) of the layer, with dimension $1 \times D$, is projected onto the transposed basis matrix $B^\top \in \mathbb{R}^{D \times C}$ to compute the logits, resulting in a logit vector of dimension $1 \times C$.}
    \label{fig:basis}
\end{figure*}

\subsection{Architecture Overview}
\label{archi}

\begin{figure*}[t]
    \centering

    \begin{minipage}[t]{0.49\textwidth}
        \centering
        \includegraphics[width=\linewidth]{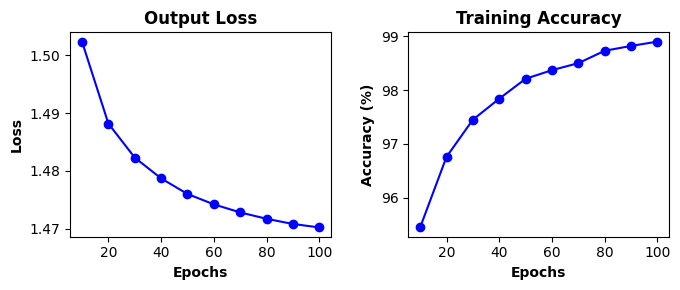}
        \caption{Results of proposed model on MNIST: (a) training loss curve showing loss decreasing after every epoch, (b) training accuracy plot showing stable learning.}
        \label{fig:mnist}
    \end{minipage}
    \hfill
    \begin{minipage}[t]{0.49\textwidth}
        \centering
        \includegraphics[width=\linewidth]{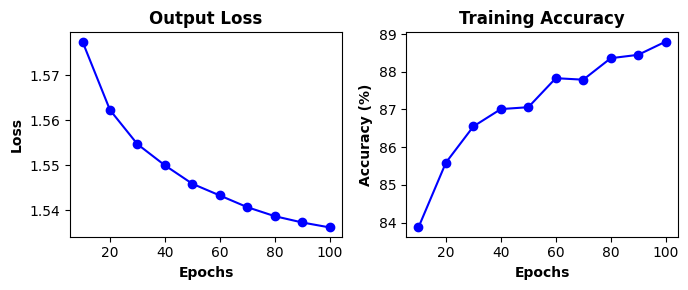}
        \caption{Results of proposed model on Fashion-MNIST: (a) training loss curve showing loss decreasing after every epoch, (b) training accuracy plot showing fluctuations during training.}
        \label{fig:fashion}
    \end{minipage}

\end{figure*}

The proposed architecture is depicted in \autoref{fig:model_arch} and is organized into three sequential stages: (i) Poisson rate encoding, (ii) Spiking neural network processing, and (iii) Local learning and neural synchronization.

The \(I\)-dimensional input is presented to the input layer and forwarded through \(n\) hidden layers. Each hidden layer is denoted by
\(\ell \in \{1, 2, ..., n\}\).
Layer \(\ell\) contains \(D^{(\ell)}\) neurons, and finally to the output layer with one neuron per class (\(C\) total). At each layer, a neuron that fires during this time step is marked with a check mark in the figure.

First, the batched input data is transformed using Poisson rate encoding, which converts each input sample into a sequence of input currents over \(T\) discrete time steps. As a result, each image is presented to the network repeatedly over time, allowing temporal spike-based processing

Next, the input layer receives the encoded spike signals \textit{x}, which move to the pre-activation phase. The product of \textit{x} and the corresponding synaptic weights \(W^{(\ell)}\) of the layer \(\ell\) is added to corresponding bias to obtain the resulting synaptic current. The synaptic current is added to the previous membrane potential \(v_{t-1}\) to update it. During activation, the Heaviside step function \(H(\cdot)\) triggers spike \(s_t\), at time step \(t\), using the updated membrane potential \(v_t\). After processing an input through all \(T\) time-steps, it emits spike signals \(s\). The average number of spikes called the spike count rate \(SC^{(\ell)}\) is used to project on basis matrix \(B^{(\ell)}\) to get local logit vector \(z^{(\ell)}\).
The spike signals \(s\) from the \(\ell^{th}\) layer is input to the \((\ell+1)^{th}\) layer.


The "Local Synchronized Weight Correction" functional block is attached to every hidden layer as shown in \autoref{fig:model_arch}. Once a hidden layer's Leaky-Integrate-Fire (LIF) neurons have fired across the time steps, its spike count rate \(SC^{(\ell)}\) is sent into this block, where it is projected onto the layer's own fixed basis matrix \(B^{(\ell)}\) to produce a local logit vector \(z^{(\ell)}\). At each hidden layer, \(z^{(\ell)}\) is compared to the target label through cross-entropy loss and a surrogate derivative. The result is fed directly into the layer's own "Weight \& Bias Update" step. The update process does not depend on activities or errors from any other layer.

This architecture, in \autoref{fig:model_arch}, depicts \(n\) independent, parallel local weight matrix update because every hidden layer has self-contained pipeline (spike count, projection, basis matrix, loss, weight matrix update). The output layer follows a different, simpler path since its spike count rate already has one value per class, it is used directly as the logit vector, with no basis matrix projection needed, and its weights are updated through the same cross-entropy and surrogate-derivative procedure.

\subsection{Basis Matrix Formation and Synchronicity Across Layers}

In \autoref{fig:basis}, each hidden layer \(\ell\) containing \(D^{(\ell)}\) neurons is assigned a fixed basis matrix
\(B^{(\ell)} \in \mathbb{R}^{D^{(\ell)} \times C}\), where \(C\) is the number of classes.
The basis matrix at each layer is made of fixed, predefined vectors generated using either sinusoidal frequency functions or random initialization.
The \(c^{th}\) row of \(B^{(\ell)}\) is a fixed basis vector assigned to class \(c\), defined over the neuron index \(i\) using a class-specific spatial frequency \(f_c\) (for sinusoidal):
\(
B_c^{(\ell)}(i) = \sin\!\left(2\pi f_c\, i\right).
\)
Stacking the \(C\) basis vectors forms \(B^{(\ell)}\). Since same set of class frequencies \({f_1, ..., f_C}\) is used at every hidden layer, every layer gives an identical class-specific target pattern causing synchronization across layers locally.
Because these vectors are fixed rather than learned, every class is represented in a separable form in the spike-count space from the start of training.
Given the  $\ell^{th}$ layer's spike count rate vector \(SC^{(\ell)}\), the layer's logits are computed as:
\[z^{(\ell)} = SC^{(\ell)} (B^{(\ell)})^\top\]
This linear projection turns raw spike activity into a local, class-comparable output at $\ell^{th}$ hidden layer. This explicit per-layer logit vector can be evaluated with a standard classification loss.
The output layer is the one exception: since its spike count rate already has one entry per class, it is used as logit
directly, with no basis matrix required.

\subsection{Local Learning Signal and Weight Update}

In layer \(\ell\), cross-entropy loss \(\mathcal{L^{(\ell)}}\) is computed from the logit \(z^{(\ell)}\) and the target vector \(y\). Differentiating \(\mathcal{L^{(\ell)}}\) with respect to \(z^{(\ell)}\) gives the class-level learning signal \(\mathrm{\delta}^{(\ell)}B^{(\ell)}\). 
The fast-sigmoid surrogate derivative \(\phi'(\cdot)\) is based on membrane potential \(v_t^{(\ell)}\) and neuron threshold \(\vartheta\). The surrogate derivative, class-level learning signal and previous layer's synaptic spike activity \(s_t^{(\ell-1)}\), at time step \(t\), are used to update the layer's weights:
\[
\Delta w^{(\ell)}
= -\eta \sum_{t=1}^{T}
\Big[\, \big(\delta^{(\ell)} B^{(\ell)}\big) \odot \phi'\!\big(v_t^{(\ell)} - \vartheta\big) \Big]^{\top}
s_t^{(\ell-1)},
\]
\[
W^{(\ell)} 
\leftarrow 
W^{(\ell)}  + \Delta w^{(\ell)}
\]
Biases are updated the same way. The \(\eta\) is the learning rate.
Since \(\mathrm{\delta}^{(\ell)}\) computed from the layer's own basis-matrix logits rather than gradient from later layers, the update does not need to store activations allowing the layers to update independently,
unlike BPTT.



\subsection{Computational Complexity}

In an SNN network with \(L\) layers, \(n\) neurons per layer, and \(T\) time steps, BPTT has time complexity \(\mathcal{O}(LTn^2)\) and memory complexity \(\mathcal{O}(LTn)\) \cite{apolinario2025tess}. In the architecture, memory usage is reduced to \(\mathcal{O}(Tn_{max})\). The proposed architecture does not propagate gradients backward through layers; memory usage is reduced to \(\mathcal{O}(Tn_{max})\), where \(n_{max}\) is the maximum neurons in any layer, and computational complexity becomes \(\mathcal{O}(LTCn)\), where \(C\) is the number of classes. This improvement in complexity makes SNN  scalable contrast BPTT-based training in resource-constrained devices.

\section{Experimental Results}
\label{experiment}

\begin{table*}[t]
\centering
\caption{Best performance of Proposed SNN model with corresponding parameters across different Datasets.}
\label{tab:result2}
\resizebox{\linewidth}{!}{
\begin{tabular}{|c|c|c|c|c|c|c|}
\hline
Dataset & \# Hidden layers & \# Neurons & \# Parameters & Learning rate & Basis type & Accuracy (\%) \\ \hline
MNIST         & 3 & 1024-256-64      & 1,083,338 & 0.1  & Periodic   & 97.78 \\ \hline
Fashion-MNIST & 2 & 2048-1024        & 3,716,106 & 0.1  & Sinusoidal & 88.06 \\ \hline
EMNIST        & 4 & 1024-512-256-128 & 1,498,927 & 0.05 & Sinusoidal & 73.62 \\ \hline
\end{tabular}
}
\end{table*}

\begin{table*}[t]
\centering
\small

\begin{minipage}[t]{0.49\textwidth}
\centering
\caption{Ablation study of the proposed SNN architecture on the MNIST dataset.}
\label{tab:abla1}
\setlength{\tabcolsep}{7pt}
\begin{tabular}{|c|c|c|c|}
\hline
\begin{tabular}[c]{@{}c@{}}\# Hidden Layers\\ (\# Neurons)\end{tabular} &
Basis Type &
\begin{tabular}[c]{@{}c@{}}Learning\\ Rate\end{tabular} &
Accuracy (\%) \\ \hline

\multirow{4}{*}{\begin{tabular}[c]{@{}c@{}}1\\ (2048)\end{tabular}}
& Sinusoidal & 0.05 & 97.16 \\ \cline{2-4}
& Sinusoidal & 0.1 & 97.53 \\ \cline{2-4}
& Random & 0.05 & 97.22 \\ \cline{2-4}
& Random & 0.1 & 97.63 \\ \hline

\multirow{4}{*}{\begin{tabular}[c]{@{}c@{}}3\\ (1024-256-64)\end{tabular}}
& Sinusoidal & 0.05 & 97.20 \\ \cline{2-4}
& Sinusoidal & 0.1 & 97.61 \\ \cline{2-4}
& Random & 0.05 & 97.40 \\ \cline{2-4}
& Random & 0.1 & 97.78 \\ \hline
\end{tabular}
\end{minipage}
\hfill
\begin{minipage}[t]{0.49\textwidth}
\centering
\caption{Ablation study of the proposed SNN architecture on the Fashion-MNIST dataset.}
\label{tab:abla2}
\setlength{\tabcolsep}{5pt}
\begin{tabular}{|c|c|c|c|}
\hline
\begin{tabular}[c]{@{}c@{}}\# Hidden Layers\\ (\# Neurons)\end{tabular} &
Basis Type &
\begin{tabular}[c]{@{}c@{}}Learning\\ Rate\end{tabular} &
Accuracy (\%) \\ \hline

\multirow{4}{*}{\begin{tabular}[c]{@{}c@{}}2\\ (2048-1024)\end{tabular}}
& Sinusoidal & 0.05 & 86.44 \\ \cline{2-4}
& Sinusoidal & 0.1 & 88.06 \\ \cline{2-4}
& Random & 0.05 & 87.38 \\ \cline{2-4}
& Random & 0.1 & 87.74 \\ \hline

\multirow{4}{*}{\begin{tabular}[c]{@{}c@{}}5\\ (1024-512-256-64-32)\end{tabular}}
& Sinusoidal & 0.05 & 86.61 \\ \cline{2-4}
& Sinusoidal & 0.1 & 86.12 \\ \cline{2-4}
& Random & 0.05 & 86.26 \\ \cline{2-4}
& Random & 0.1 & 86.57 \\ \hline
\end{tabular}
\end{minipage}

\end{table*}

\begin{table*}[t]
\caption{Comparative analysis of the Proposed SNN architecture and Method against Existing State-of-the-Art SNN approaches on MNIST Dataset.}
\label{tab:compare}
\resizebox{\linewidth}{!}{%
\begin{tabular}{|c|c|c|c|c|c|c|}
\hline
Paper & Training Method & \begin{tabular}[c]{@{}c@{}}\# Hidden Layers\\ (\# Neurons)\end{tabular} & Time-Steps & Epochs & Local Learning & Accuracy (\%) \\ \hline
CoLaNET \cite{kiselev2025colanetspikingneural} & Supervised STDP with dopamine-modulated plasticity & \begin{tabular}[c]{@{}c@{}}1 (150)\end{tabular} & 20 & - & - & 96.87 \\ \hline
CDNA-SNN \cite{10428047} & Supervised STDP with CDNA-Weighted STDP & 2 (55-38) & 300 & 100 & Yes (unsupervised) & 98.91 \\ \hline
HM2-BP \cite{jin2019hybridmacromicrolevelbackpropagation} &  Hybrid macro/micro level backpropagation & 1 (800) & 400 & 100 & No (supervised) & 98.93 \\ \hline
STDBP \cite{DBLP:journals/tnn/ZhangWWBAZMQCCL22}  & Spike-Timing-Dependent Backpropagation & 1 (800) & - & 150 & No (supervised) & 98.50 \\ \hline
BS4NN \cite{10.1007/s11063-021-10680-x} &  BPTT + Binary synaptic weights & 1 (600) & 256 & 500 & No (supervised) & 97.00 \\ \hline
PSAC \cite{Nazari2025} & STDP with Reinforcement learning & 1 (7010) & - & 6 & Yes (unsupervised) & 98.10 \\ \hline


\textbf{This work} & \textbf{Local Learning + NAS} & \textbf{3 (1024-256-64)} & \textbf{20} & \textbf{100} & \textbf{Yes (supervised)} & \textbf{97.78} \\ \hline
\end{tabular}
}
\end{table*}

\begin{table*}[t]
\caption{Comparative analysis of the Proposed SNN architecture and Method against Existing State-of-the-Art SNN approaches on Fashion-MNIST Dataset.}
\label{tab:compare2}
\resizebox{\linewidth}{!}{%
\begin{tabular}{|c|c|c|c|c|c|c|}
\hline
Paper & Training Method & \begin{tabular}[c]{@{}c@{}}\# Hidden Layers\\ (\# Neurons)\end{tabular} & Time-Steps & Epochs & Local Learning & Accuracy (\%) \\ \hline
CDNA-SNN \cite{10428047} & Supervised STDP with CDNA-Weighted STDP & 2 (53-46) & 300 & 100 & Yes (unsupervised) & 90.12 \\ \hline
STDBP \cite{DBLP:journals/tnn/ZhangWWBAZMQCCL22}  & Spike-Timing-Dependent Backpropagation & 1 (1000) & - & 150 & No (supervised) & 88.10 \\ \hline
BS4NN \cite{10.1007/s11063-021-10680-x} &  BPTT + Binary synaptic weights & 1 (1000) & 256 & 500 & No (supervised) & 87.30 \\ \hline
\textbf{This work} & \textbf{Local Learning + NAS} & \textbf{2 (2048-1024)} & \textbf{20} & \textbf{100} & \textbf{Yes (supervised)} & \textbf{88.06} \\ \hline
\end{tabular}
}
\end{table*}

The performance of the proposed model has been evaluated based on several design choices to achieve an optimal balance between accuracy, computational cost, and simplicity of the architecture. Three benchmark datasets: MNIST \cite{726791}, Fashion-MNIST \cite{xiao2017fashionmnistnovelimagedataset}, and EMNIST \cite{7966217}
are chosen for experiments.
The models were implemented using PyTorch framework on a system with configurations of $8$ GB RAM and $4$ GB GPU. 

For MNIST and Fashion-MNIST, the number of time-steps and the number of training epochs were $20$ and $100$ respectively. The key experimental parameters are as follows: (i) the fixed basis matrix constructed with sinusoidal function in frequencies range from $20$ to $2000$, (ii) the firing threshold is set to $\vartheta = 0.5$, (iii) the batch size is $256$, and (iv) input pixel values are normalized from $[0,255]$ to $[0.0,1.0]$.

\autoref{tab:result2} shows the best results achieved by the proposed model on the mentioned datasets, along with corresponding model configurations. In order to analyze the performance of the model, an ablation study was conducted across the datasets by varying parameters, basis function type and learning rate. The corresponding experimental results are presented in Tables \ref{tab:abla1} and \ref{tab:abla2} for MNIST and Fashion-MNIST, respectively.

$97.63\%$ classification accuracy is recorded for the proposed model on the MNIST dataset, with one hidden layer consists of $2048$ neurons. In contrast, many of existing models rely on backpropagation, deeper network architectures, or significantly larger parameter counts.
In the proposed model with three hidden layers and comparatively lesser neurons, it gives $97.78\%$ accuracy.
These demonstrate the robustness of the model. \autoref{fig:mnist} (a) shows the training loss curve
and (b) demonstrates the training accuracy plot. 

In order to test resilient against input noise, the model is tested with Gaussian and Salt-and-Pepper noises into test data set. Adding Gaussian noise into the test data of MNIST with standard deviation values of $0.1$ and $0.2$ the model achieved accuracies of $96.85\%$ and $94.61\%$ respectively while introducing Salt-and-Pepper noise into the MNIST test data at noise densities of $0.1$ and $0.2$, the model achieved accuracies of $92.22\%$ and $84.64\%$, respectively.

The model was further evaluated on Fashion-MNIST dataset with two-layered architecture having $2048$ and $1024$ neurons, respectively. It achieved $88.06\%$ classification accuracy with this simple network architecture.
\autoref{fig:fashion} (a) and (b) show training loss
and training accuracy plot, respectively.

The model was also trained and tested on the more challenging EMNIST balanced dataset with four hidden layers $(1024-512-256-128)$ achieving $73.62\%$ test accuracy.

In the \autoref{tab:compare} and \autoref{tab:compare2}, the results of proposed model are compared with the existing State-of-the-Art works, showing the proposed method requires, comparatively low time-steps and parameters to achieve comparable performance.

\section{Conclusion}
\label{conclusion}

This paper presents a local trainable Spiking Neural Network architecture integrating Neural Activity Synchronization using fixed basis matrices. In this architecture, no gradient backpropagation is required. It significantly reduces the computational and memory overhead achieving $97.78\%$ accuracy on MNIST dataset. The model shows robustness against Gaussian and Salt-and-Pepper noise supporting its practical applicability. Future direction will focus on extending the approach for continual learning and deploying it on resource-constraint hardware platform.



\printbibliography

\end{document}